\documentclass[11pt,a4paper]{article}
\usepackage[hyperref]{acl2019}
\usepackage{times}
\usepackage{latexsym}
\usepackage[pdftex]{graphicx}
\usepackage{booktabs}

\usepackage{amsmath}
\usepackage{amsfonts}  

\usepackage{url}

\aclfinalcopy{}

\newcommand{\indic}[1]{\mathbb{I}_{\left [ #1 \right ]}}
\newcommand{\pic}{\pi^\textsc{c}}
\newcommand{\piexp}{\pi^\textsc{exp3.s}}
\newcommand{\wacl}{\omega}
\newcommand{\wc}{\omega}

\title{Task Selection Policies for Multitask Learning}

\author{John Glover \and Chris Hokamp \\
  Aylien \\
  \texttt{<first-name>@aylien.com}
}

\date{}

\begin{document}
\maketitle

\begin{abstract}
    One of the questions that arises when designing models that learn to
    solve multiple tasks simultaneously is how much of the available
    training budget should be devoted to each individual task.
    We refer to any formalized approach to addressing this problem (learned or otherwise)
    as a \textit{task selection policy}.
    In this work we provide an empirical evaluation of the performance of
    some common task selection policies in a synthetic bandit-style setting,
    as well as on the GLUE benchmark for natural language understanding.
    We connect task selection policy learning to existing work on automated
    curriculum learning and off-policy evaluation,
    and suggest a method based on counterfactual estimation that leads to
    improved model performance in our experimental settings.
\end{abstract}

\section{Introduction}
Recent work on language understanding has demonstrated the
effectiveness of pretraining neural networks
on large corpora using unsupervised objectives such as language modelling,
and then fine-tuning the resulting models on downstream target
tasks~\cite{dai2015,peters2018,howard2018a,radford2018,devlin2018}.
This approach has produced new state-of-the-art results on a variety of
popular benchmark datasets, such as the SQuAD question answering
dataset~\cite{rajpurkar2016}, and the General Language Understanding Evaluation (GLUE) benchmark
for sentence (and sentence pair) classification~\cite{wang2018a}.
Notably, these approaches typically fine-tune a full copy of the pretrained model on
each target task individually, effectively multiplying the number of parameters that
must be trained and stored by the number of tasks, and ruling out any potential
performance improvements that may arise from sharing information between related tasks.

An alternative approach is to do \textit{Multitask Learning} (MTL)~\cite{caruana1997},
where a model is learned that shares some number of parameters across all tasks.
BERT is a recent example of the benefits of this approach --- in the pretraining stage it is
trained on two tasks simultaneously: masked language modelling (predicting missing tokens in the input),
and next sentence prediction (predicting whether two sentences are consecutive or not).
However, successfully applying MTL to a particular problem is not necessarily straightforward,
and depends on resolving many questions that do not arise in the single task setting, such as:
\begin{enumerate}
    \item In what settings is MTL effective?
        Can we detect and mitigate \textit{negative transfer} (where performance on some subset
        of tasks decreases when trained in an MTL setting)?
    \item Which parameters of the model should be shared between tasks, and which
        should be task specific?
    \item Should the model look at all of the data from all of the tasks?
        Should all of the training examples be weighted equally in the loss term?
    \item How much of the available training budget should be spent on each individual task?
\end{enumerate}

This work focuses on question 4, providing an empirical evaluation of the performance of different
policies for selecting how much to sample from each task in two experimental settings;
firstly on a novel bandit-style task, and secondly when fine-tuning a pretrained language model
(BERT) on the GLUE benchmark.
GLUE represents an interesting challenge as it evaluates the performance of a single model
across multiple tasks, such as textual entailment and question answering, with the size of the available training data for different tasks spanning multiple orders of magnitude.
We find that policies based on common heuristics such as sampling tasks uniformly at
random~\cite{wang2018a,subramanian2018} or proportionally to their size~\cite{phang2018}
are not able to match the performance of fine-tuning models for each individual task.

We also show that this problem can also be viewed through the lens of \textit{curriculum learning}.
We evaluate a previous method for automated curriculum learning~\cite{graves2017},
but find that on our tasks it does not perform significantly better than a random policy.
However, when learning a policy using counterfactual estimation~\cite{bottou2012}, we are able to
approach performance parity with the task-specific models.


\section{Related Work}
Multitask learning has been studied extensively,
and can be motivated as a means of inductive bias learning~\cite{caruana1993,baxter2000},
representation learning~\cite{argyriou2007,misra2016},
or as a form of learning to learn~\cite{baxter1997,thrun1998,heskes2000,lawrence2004}.
In the context of natural language processing, MTL has been used
to improve tasks such as semantic role labeling~\cite{collobert2008,strubell2018},
and machine translation~\cite{luong2016a,firat2016,johnson2016a,hokamp2018},
and to learn general purpose sentence representations~\cite{subramanian2018}.
One of the best performing models on the GLUE benchmark at the time of publication~\cite{liu2019b}
combines MTL pretraining (BERT) with MTL fine-tuning on the GLUE tasks, although the model also
incorporates non-trivial task-specific components and additional training using
knowledge distillation~\cite{hinton2015}.

However, MTL in NLP is not always successful --- the GLUE baseline MTL models are significantly
worse than the single task models~\cite{wang2018a}, \citealp{alonso2016} only find significant improvements
with MTL in one of five tasks evaluated,
and the multitask model in \citealp{mccann2018} does not quite reach the performance of the
same model trained on each task individually.

One way to approach the question of how much training budget to spend on each task
is to view this as a curriculum learning problem~\cite{elman1993,bengio2009}.
\citealp{mccann2018} show the importance of using the correct curriculum to train their multitask model.
\citealp{graves2017} treat curriculum learning as an adversarial multi-armed bandit
problem~\cite{auer2002,bubeck2012a}, and show that the learned policies
can improve the performance on language modeling and the bAbI dataset tasks~\cite{weston2015}.
Probably the most similar recent work to ours is AutoSeM~\cite{guo2019}, where they first learn
to select useful auxiliary tasks, then learn an MTL curriculum over them using Bayesian optimization.
Our work differs in the methods used (we use counterfactual estimation to learn the curriculum),
as well as in the learning objective --- \citealp{guo2019} focus on the improving a \textit{single}
target task by incorporating auxiliary tasks, where as we seek to jointly maximize the performance
of a set of target tasks.

Counterfactual estimation~\cite{bottou2012} tries to answer the question ``what would have happened if
an agent had taken different actions?'', and so is closely related to the problem of
off-policy evaluation in the context of reinforcement learning\footnote{Counterfactual estimation has
also been studied under the names ``counterfactual reasoning'' and ``learning from logged bandit feedback''.}.
This setting presents an added set of challenges to on-policy evaluation, as an agent only has partial
feedback from the environment, and it is assumed that collecting additional feedback is either not
possible or prohibitively expensive.
Typically approaches to off-policy evaluation are based on either modeling the environment dynamics and reward, using importance sampling, or a combination of the two~\cite{precup2000,peshkin2002,dudik2011,swaminathan2015,jiang2015}.


\section{Task Selection Policies}
We consider the case where we have a set of $N$ learning tasks.
Each task is a distribution $D_k$ ($k \in \{1,\ldots, N\}$) over samples $\mathbf{x}$
from the input space $\mathcal{X}$.
In the supervised case for example, $\mathbf{x}$ may be composed of $(x, y)$ pairs,
where $x$ is an input example or sequence and $y$ is a target label or sequence.
These individual samples $\mathbf{x}$ are typically grouped into batches of multiple samples from the
same task, we also refer to these batches as samples for convenience.

A model over $\mathcal{X}$ has parameters $\theta$.
A loss function $L_k(\mathbf{x}, \theta)$ is defined for each task, with the expected
loss for the $k^{th}$ task given by
\begin{equation}
    \mathcal{L}_k(\theta) = \mathop{\mathbb{E}}_{\mathbf{x} \sim D_k} \left[ L_k(\mathbf{x}, \theta) \right]
\end{equation}
The objective is to maximize the performance on all tasks, or to minimize the total
loss:
\begin{equation}
    \mathcal{L(\theta)} = \frac{1}{N} \sum_{k = 1}^{N} \mathcal{L}_k(\theta)
    \label{eqn:loss}
\end{equation}

We assume that we have a fixed training budget of $T$ steps.
At each step $t \in \{1,\ldots,T\}$, a model
samples task (or \textit{action}) $k$ according to a distribution defined by
some \textit{task selection policy} $\pi$,
then processes the resulting sample and observes loss $L_k(\mathbf{x}, \theta)$.
The probability of selecting a particular action at time $t$ is given by $\pi_t(k)$.
We use $\theta_\pi$ to denote a model with parameters $\theta$ that uses policy $\pi$,
and $\omega$ to refer to any additional parameters that $\pi$ itself may have.
In general, $\omega$ may be a function of the sequence of the observed
samples, losses and model parameters in the training history at time $t$.
In this work we study different methods for specifying or learning $\pi$, and
how these approaches influence $\mathcal{L(\theta_\pi)}$.
We are primarily interested in cases where the model is a large neural network,
and $\{D_1,\ldots,D_k\}$ is a large dataset, so in general
we seek methods that avoid evaluating $\mathcal{L(\theta_\pi)}$ for many different values of $\pi$
as this is computationally slow and expensive.

\subsection{Baseline (Heuristic) Policies}
A common task selection policy is to sample from all tasks uniformly at random:
\begin{equation}
\pi_t^{\textsc{random}}(k) = \frac{1}{N}
\end{equation}
This policy has been shown to be a strong baseline in previous work on curriculum learning~\cite{graves2017}.
Another common heuristic that we evaluate in this work is to sample tasks proportionally to their dataset size~\cite{phang2018}:
\begin{equation}
    \pi^{\textsc{task size}}_{t}(k) = \frac{|D^k|}{\sum_{j = 1}^{N}|D^j|} \\
\end{equation}

\subsection{Learning Policies Using Automated Curriculum Learning}
We also evaluate the approach to automated curriculum learning introduced in \citealp{graves2017},
which is briefly described here for reference.
A curriculum of $N$ tasks can be viewed as an $N$-armed bandit.
At each round (time step) $t$ an agent uses a policy $\pi$ to select an action
and sees a reward $r^\textsc{acl}_t$.
The goal is to create a policy that maximizes the total reward from the bandit.

\citealp{graves2017} use the Exp3.S algorithm~\cite{auer2002} for their policy,
which is an adaptation of the Exp3 algorithm to the non-stationary setting.
Exp3 tries to minimize the regret with respect to a single best arm in hindsight, assuming an
adversarial setting in which the distribution of rewards over arms can change
at every time step.
It does this by using importance sampled rewards to update a set of weights $\wacl$, then
acting stochastically according to a distribution based on these weights
(with additional hyperparameters $\eta$ and $\epsilon$):
\begin{equation}
\begin{split}
    \wacl_{t,k} &= \log \Big [ (1 - \alpha_t) \exp\left \{\wacl_{t-1,k} + \eta {\tilde r}_{t-1,k} \right \} \\
                &+ \frac{\alpha_t}{N-1}
                   \sum_{j \ne k} \exp \left \{\wacl_{t-1,j} + \eta {\tilde r}_{t-1,j} \right\} \Big ] \\
    \wacl_{1,k} &= 0 \quad \alpha_t = t^{-1} \\
    {\tilde r}_{t,k} &= \frac{r^\textsc{acl}_t \indic{a_t = k}}{\rho_{t,k}} \\
    \rho_{t,k} &= \frac{e^{\wacl_{t,k}}}{\sum_{j=1}^N  e^{\wacl_{t,j}}} \\
    \piexp_t(k) &= (1 - \epsilon) \rho_{t,k} + \frac{\epsilon}{N} \\
\end{split}
\end{equation}

\subsubsection{Automated Curriculum Learning Reward Function}
In \citealp{graves2017} the underlying hypothesis is that the policy should produce a syllabus that focuses on tasks in order
of increasing difficulty.
This lead the authors to design and evaluate several different ways of encoding measures of the rate at which learning progresses
into a sample-level reward function. They find that the progress signal that they call \textit{prediction gain}
generally leads to the best performance across the tasks that they evaluated, and so that is the raw reward
signal that we consider with $\piexp$ here. Prediction gain is defined as the change in loss
before and after training on a sample $\mathbf{x}$ (ie., after a gradient update):
$L_k(\mathbf{x}, \theta_{\piexp}) - L_k(\mathbf{x}, \theta'_{\piexp})$.

When computing $\piexp$ we also follow the reward scaling process described in
\citealp{graves2017}. Reservoir sampling is used to maintain a representative sample of the unscaled
reward history up to time $t$, and from this we compute the 20$^{\text{th}}$ and 80$^{\text{th}}$
percentiles as $q^{20}_t$ and $q^{80}_t$ respectively.
The unscaled reward $\hat{r}_t$ is then mapped to the interval $[-1, 1]$:
\begin{equation}
    r^\textsc{acl}_t=
\begin{cases}
    -1 & \text{if } \hat{r}_t < q^{20}_t \\
    1  & \text{if } \hat{r}_t > q^{80}_t \\
    \frac{2 (\hat{r}_t - q^{20}_t)}{q^{80}_t - q^{20}_t} - 1 & \text{otherwise} \\
\end{cases}
\end{equation}

\subsection{Learning Policies Using Counterfactual Estimation}
Automated curriculum learning using Exp3.S describes a method for online learning of task selection
policies from bandit-style feedback.
However in the context of MTL, the ability to learn online is typically not required ---
in many situations we have the ability to run multiple variations of a given experiment,
and can potentially use the results of earlier training runs in order to improve our policies.
Learning task selection policies can therefore be viewed as a problem of \textit{counterfactual estimation},
where the goal is to use old policy data to improve on that policy without further
interaction with the environment.

Most approaches to counterfactual estimation either model the reward generating process,
use importance sampling to correct for the changes introduced by the new policy,
or combine the two~\cite{dudik2011}.
In the first case, the task initially reduces to a supervised regression problem, followed by a
process of policy optimization using the reward model as the target reward that should be
maximized.
However, defining a suitable MTL sample-level reward function is non-trivial.
In this work we are interested in maximizing the average performance
of all training tasks, where performance is measured \textit{at the end of the training run}.
In problems with large models and/or
datasets, this could take anywhere from hours to weeks to complete, and so the most obvious reward
signal (final average performance) is very sparse and difficult to optimize as the time that this reward is observed at is
potentially very delayed from the time when an action must be taken.
We therefore tend to rely on reward signals that are defined in response to each action taken
by the policy, with the hope that they correlate well with the desired global metric that
we care about.
We discuss the specific variant of the sample-level reward used in this work further in
Section~\ref{sec:cf_reward} below.

As we are dealing with a surrogate reward signal, and previous work on reward modelling has found
that this approach may not generalise well even with exact rewards~\cite{beygelzimer2009}, we instead focus on the
counterfactual estimation methods that are based on importance sampling.
We use the available data in a two-step process to create task selection policies:
\begin{enumerate}
    \item Create an estimator to evaluate a given policy, using some set of logged policy probabilities,
        decisions and resulting rewards from some initial training run(s).
    \item Create a new policy that maximises the expected reward according to this counterfactual estimator.
\end{enumerate}
These steps are described in detail in Sections~\ref{sec:cf} and~\ref{sec:cf_policy} respectively.

\subsubsection{Counterfactual Estimation}\label{sec:cf}
At each time step in a learning process, an MTL model selects a task to sample from using a
policy $\pi$, and in response experiences a sample $\mathbf{x}_t$ and reward signal $r_t(\mathbf{x}_t)$.
The probability of selecting each $\mathbf{x}$ is given by a distribution $P(\mathbf{x} | \pi)$.
The value of $\pi$ is the expected reward obtained when selecting tasks using that policy:
\begin{equation}
    V(\pi) = \mathop{\mathbb{E}}_{\mathbf{x} \sim P(\mathbf{x} | \pi)} \left[ r(\mathbf{x}) \right]
           = \sum_{\mathbf{x}} P(\mathbf{x} | \pi) r(\mathbf{x})
\end{equation}
$V(\pi)$ can be estimated by sampling trajectories (or rollouts) from
$P(\mathbf{x} | \pi)$\footnote{To simplify our notation it is assumed that we are just using a single
rollout to compute $\hat{V}(\pi)$, but multiple rollouts could be used.}:
\begin{equation}
    \hat{V}(\pi) = \frac{1}{T} \sum_{t = 1}^{T} r_t(\mathbf{x}_t)
\end{equation}

In counterfactual estimation, the goal is to use rollouts from $\pi$ to approximate
the value of a different policy $V(\pic)$, under the assumption that we cannot sample directly from
$P(\mathbf{x} | \pic)$.
We also cannot evaluate the reward function for samples that are selected using $\pic$ but not $\pi$.
One way to estimate $V(\pic)$ is to use importance sampling~\cite{rosenbaum1983}:
\begin{equation}
\begin{split}
    V(\pic) &= \sum_{\mathbf{x}} P(\mathbf{x} | \pic) r(\mathbf{x}) \\
            &= \sum_{\mathbf{x}} P(\mathbf{x} | \pic) \frac{P(\mathbf{x} | \pi)}{P(\mathbf{x} | \pi)} r(\mathbf{x}) \\
            &= \sum_{\mathbf{x}} P(\mathbf{x} | \pi) \frac{P(\mathbf{x} | \pic)}{P(\mathbf{x} | \pi)} r(\mathbf{x}) \\
            &= \mathop{\mathbb{E}}_{\mathbf{x} \sim P(\mathbf{x} | \pi)} \left[
                \frac{P(\mathbf{x} | \pic)}{P(\mathbf{x} | \pi)} r(\mathbf{x})
            \right]
\end{split}
\end{equation}
We can therefore use Monte Carlo to approximate $V(\pic)$ while only relying on samples from $P(\mathbf{x} | \pi)$:
\begin{equation}
    \hat{V}_\text{IS}(\pic) = \frac{1}{T} \sum_{t = 1}^{T}
        \frac{P(\mathbf{x}_t | \pic)}{P(\mathbf{x}_t | \pi)} r_t(\mathbf{x}_t)
\end{equation}
The importance sampling estimator (also known as the inverse propensity score estimator) is unbiased,
and is defined as long as $\pi$ has support everywhere that $\pic$ does, or
in other words if $P(\mathbf{x} | \pic) > 0 \implies P(\mathbf{x} | \pi) > 0$.
In practise this is not a significant limitation, as we generally have full control over $\pi$, and so can
ensure that it always assigns some probability mass to each task.

However, the importance sampling estimator is known to suffer from high variance~\cite{bottou2012,joachims2018}.
This problem is particularly noticeable in regions of the input space that are not well covered
by the sampled policy $\pi$ --- if $P(\mathbf{x}_t | \pi)$ is very low for a particular sample, then the
importance weight $\frac{P(\mathbf{x}_t | \pic)}{P(\mathbf{x}_t | \pi)}$ will be high (leading to
inaccurate estimates) unless the reward for this sample is very low.
Different estimators have been proposed that reduce this variance in some way, generally at the expense of
adding some bias~\cite{dudik2011,bottou2012}, but there is no single estimator that works best in all
situations~\cite{nedelec2017}.

In this work we use the \textit{weighted importance sampling} estimator~\cite{rubinstein1981}
to reduce the variance of $\hat{V}_\text{IS}$, as it has been shown to work well in a variety
of settings~\cite{mahmood2014,swaminathan2015a,nedelec2017}:
\begin{equation}\label{eqn:wis}
\begin{split}
    Z &= \frac{1}{T} \sum_{t = 1}^{T} \frac{P(\mathbf{x}_t | \pic)}{P(\mathbf{x}_t | \pi)} \\
    \hat{V}_\text{WIS}(\pic) &= \frac{\hat{V_\text{IS}}(\pic)}{Z}
\end{split}
\end{equation}
So far we have been assuming that the off-policy data is collected using a single policy $\pi$,
but these methods can also be applied to data collected by multiple logging policies~\cite{peshkin2002,agarwal2017a}.

\subsubsection{Counterfactual Estimation: Policy Improvement}\label{sec:cf_policy}

The estimators described in Section~\ref{sec:cf} can be used to evaluate arbitrary
policies, and so they can be combined with policy search algorithms to learn an
improved policy $\pi^\textsc{c}$.
In general these policies may be dynamic, but here we consider fixed stochastic policies
parameterised by a vector $\wc$, of the form:
\begin{equation}
    \pi_{\wc,t}^{\textsc{c}}(k) = \frac{e^{\wc_k}}{{\sum_{j=1}^N  e^{\wc_j}}}
\end{equation}
The learning objective is now to find an appropriate $\wc$:
\begin{equation}
    \mathop{\max}_{\wc} \hat{V}_\text{WIS}(\pi_{\wc,t}^\textsc{c})
\end{equation}
To optimize for $\wc$ we use Covariance-Matrix Adaptation Evolution Strategy (CMA-ES)~\cite{hansen2001},
as it has been shown to work well in low-dimensional parameter spaces~\cite{ha2018}.

In initial experiments, we found that the output of $\hat{V}_\text{WIS}$ was still prone to overestimating
the value of regions of the parameter space that were not well-represented in the data from the
logging policy.
In particular, it tended to assign high scores to policies with distributions that were very peaked
around the individual tasks with the largest total reward.
This is likely due to a combination of overfitting to a small amount of logging data, as well as
deficiencies with the surrogate reward signal (described in Section~\ref{sec:cf_reward}).
We therefore introduced a regularization term to the objective, limiting the parameter space to
regions where the policy retains larger entropy ($\text{H}$) values (weighted by a hyperparameter $\lambda$):
\begin{equation}
    \mathop{\max}_{\wc} \left[
        \hat{V}_\text{WIS}(\pi_{\wc,t}^\textsc{c}) + \lambda \text{H}(\pi^\textsc{c}_{\wc,t})
    \right]
\end{equation}

\subsubsection{Counterfactual Estimation: Reward}\label{sec:cf_reward}
To learn a task selection policy using counterfactual estimation we need to define a reward function $r$.
Ideally, a policy that maximises the expected value of $r$ will also minimise the average task
loss $\mathcal{L}$ at the end of training:
\begin{equation}
\mathop{\max}_{\pi} \mathop{\mathbb{E}}_{\mathbf{x} \sim P(\mathbf{x} | \pi)} \left[ r(\mathbf{x}) \right]
\approx
\mathop{\min}_{\pi} \mathcal{L}(\theta_\pi).
\end{equation}
We found that maximising the prediction gain reward proposed in~\citealp{graves2017} did not correlate
well with minimising $\mathcal{L}$ in our initial experiments, and so use a slightly different reward
formulation here.
Intuitively, as we want to maximise the average task performance, we want to incentivise spending more
time on tasks that are performing poorly at a given phase of the training process, as long as we are continuing
to make progress on those tasks. We want to penalise sampling from tasks that are not improving, as this is
a waste of effort, regardless of their overall performance.

Concretely, we assume that the sample loss for a given task $L_k$ is a negative log likelihood value.
We compare $L_k$ at time $t$ with $L_k$ at time $t - \delta_k$, which is the time at which we \textit{last sampled from the same task}.
If the difference between the two ($\Delta_L$) is negative (ie., the loss has decreased),
then the reward received by the model for this action is $1.0 - P(\mathbf{x} | \theta)$, so the model
receives a reward in the interval $[0, 1]$, with higher values for sampling from tasks that are performing
poorly.
If $\Delta_L \geq 0$, the reward is $0$.
This reward process is described in Equation~\ref{eqn:reward}.
\begin{equation}\label{eqn:reward}
\begin{split}
    \Delta_L &= L_k(\mathbf{x}_t, \theta_t) - L_k(\mathbf{x}_{t - \delta_k}, \theta_{t - \delta_k}) \\
    r_t &=
\begin{cases}
    1.0 - e^{-L_k(\mathbf{x}_t, \theta_t)} & \text{if } \Delta_L < 0 \\
    0 & \text{otherwise} \\
\end{cases}
\end{split}
\end{equation}


\section{Experiments}
To compare the different task selection policies for MTL we evaluate their performance
on two tasks: a toy bandit-style problem, and the GLUE benchmark for natural language understanding.
Further details are given in Sections~\ref{sec:bandit} and~\ref{sec:glue} respectively.

\subsection{Bandit Example}\label{sec:bandit}
Our first experiment aims to verify that our approach to learning task selection policies
works in a synthetic setting, that was designed to be a simplified environment that
still presents some of the challenges that are experienced with MTL in more realistic scenarios.
We define an MTL bandit as a bandit with $N$ arms, representing our $N$ tasks.
We assume that the goal is to sample from these arms according to some fixed oracle
distribution $\pi^\textsc{oracle}$ that is unknown to the agent interacting with
the bandit environment.
At the start of the experiment we sample $\pi^\textsc{oracle} \sim \text{Dir}(\alpha^\textsc{mtl})$,
and the same $\pi^\textsc{oracle}$ is used for all experiment runs.
Each arm has an associated score $(\text{score}_k)$, which is initially zero.
Each arm is assigned a maximum probability value ($\text{MaxP}_k$) that it can obtain in the range
$[0.5, 1.0]$, and then establishes a learning increment ($\text{LI}$) using the formula:
\begin{equation*}
\text{LI}_k = \text{MaxP} / (T * \pi^\textsc{oracle}(k))
\end{equation*}
where $k$ is the arm and $T$ is the number of steps. Each arm is also assigned a forget
increment ($\text{FI}_k$), which is a random number in the range $[0.0, 0.01]$ multiplied
by $\text{LI}_k$.
At each time step, if arm $k$ is selected, $\text{score}_k$ is incremented by $\text{LI}_k$
(and constrained to be $\leq \text{MapP}_k$), simulating some improvement on that task.
Similarly, $\text{score}_k$ is reduced by $\text{FI}_k$ (constrained to be $\geq 0$) for all tasks
at each time step, whether that task was selected or not, simulating some form of forgetting
on each task.
$\text{score}_k$ for the $k$ selected by the policy being evaluated is computed before and after
applying $\text{LI}_k$ and $\text{FI}_k$, and is used to compute rewards $r^\textsc{acl}$ and $r$
for $\piexp$ and $\pic$ respectively.

\begin{figure}[h]
    \centering
    \includegraphics[width=\linewidth]{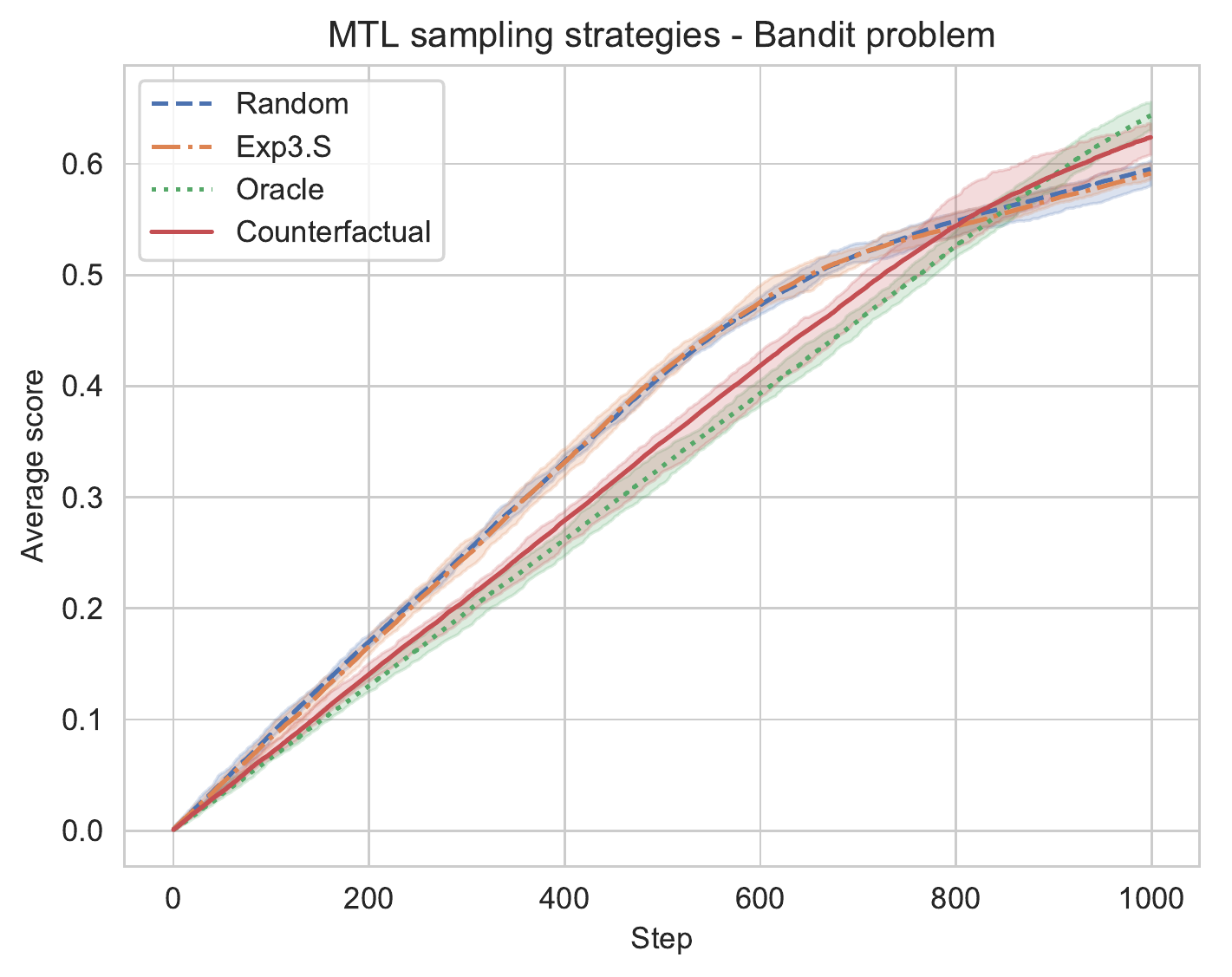}
\caption{Average score across all 8 arms over time for each policy.
Each line is the median value over 10 seeds, with shaded regions encompassing all seeds.}\label{fig:bandit}
\end{figure}

The MTL bandit creates an environment in which successful agents must learn to sample from all tasks
periodically so as not to ``forget'', but should learn that some tasks need to be sampled from
more than others in order to maximise the overall average score.
We evaluated $\pi^\textsc{random}$, $\piexp$ and $\pic$ on this task.
In \citealp{graves2017} the authors used the same hyperparameters for all experiments, and so we
use the same settings here: $\eta = 10^{-3}$, $\epsilon = 0.05$.
To set the $\pi^\textsc{c}$ entropy weight $\lambda$, we perform a grid search over
$\{0.1, 0.15, 0.2, 0.25\}$, and select the best performing value ($0.2$).
For CMA-ES we used 20 iterations with a population size of 64.
$\pi^\textsc{c}$ is computed based on 2 iterations of policy improvement, starting from a random
uniform policy (ie., $\pi^\textsc{random}$).
For the MTL bandit, we set $N = 8$ and $\alpha^\textsc{mtl} = 2.0$.
We run each policy 10 times (with different initial random seeds).
The results are shown in Figure~\ref{fig:bandit}.
None of the methods are able to fully match the oracle performance, but our counterfactual method
comes close. The random policy and Exp3.S perform similarly, both noticeably worse than the
counterfactual policy.

\subsection{GLUE}\label{sec:glue}
We evaluated the task selection policies in a more challenging and realistic environment --- the GLUE
benchmark.
GLUE performance is based on an average score across the following tasks:
CoLA (the Corpus of Linguistic Acceptability),
MNLI (Multi-Genre Natural Language Inference),
MRPC (the Microsoft Research Paraphrase Corpus),
QNLI (Question Natural Language Inference, a version of the Stanford Question Answering Dataset),
QQP (Quora Question Pairs),
RTE (Recognizing Textual Entailment),
SST-2 (the Stanford Sentiment Treebank),
STS-B (the Semantic Textual Similarity Benchmark),
and WNLI (Winograd Natural Language Inference).
We refer readers to~\citealp{wang2018a} for further details on the various tasks.

We use the same underlying model with identical hyperparameters to evaluate each policy, only varying the
policy itself and the random seed for each run.
We use the pretrained $\text{BERT}_\textsc{base}$ model, and follow a similar procedure to fine-tune on
GLUE to the one described by the BERT authors in~\citealp{devlin2018}.
We take the final hidden vector corresponding to the first input token as the representation of
the sentence (or sentence pair) for each task.
The only task-specific parameters that are used are for the output layers for each task (mapping from
the BERT hidden size to the number of output labels that task), all other model parameters are shared
across all tasks.
One departure from the details in~\citealp{devlin2018} is that we use a maximum sequence length of 256 (instead
of 512), as we don't notice a significant performance difference, and it allows us to fit one full
batch (size 16) into memory on an NVIDIA P100 GPU\@.
We use a learning rate of $2e^{-5}$, and train for 200000 steps.
To match the evaluation in~\citealp{devlin2018}, we only train on 8 of the GLUE tasks, excluding WNLI\@.
When reporting GLUE test set results, we output the majority class label for the WNLI task.
For $\piexp$ we again use $\eta = 10^{-3}$, $\epsilon = 0.05$.
$\pic$ is computed based on a single policy improvement iteration, starting from the
output of $\pi^\textsc{random}$.
We used 50 iterations of CMA-ES with a population size of 64.
We compute $\pi^\textsc{c}$ distributions for each $\lambda$ value in
$\{0.1, 0.15, 0.2\}$ and run one iteration for each policy. We then pick the
highest performing model ($\lambda = 0.15$) and use this policy for additional runs.
We run the experiment 3 times for each policy, with different random seeds each time.

\begin{figure}[h]
    \centering
    \includegraphics[width=\linewidth]{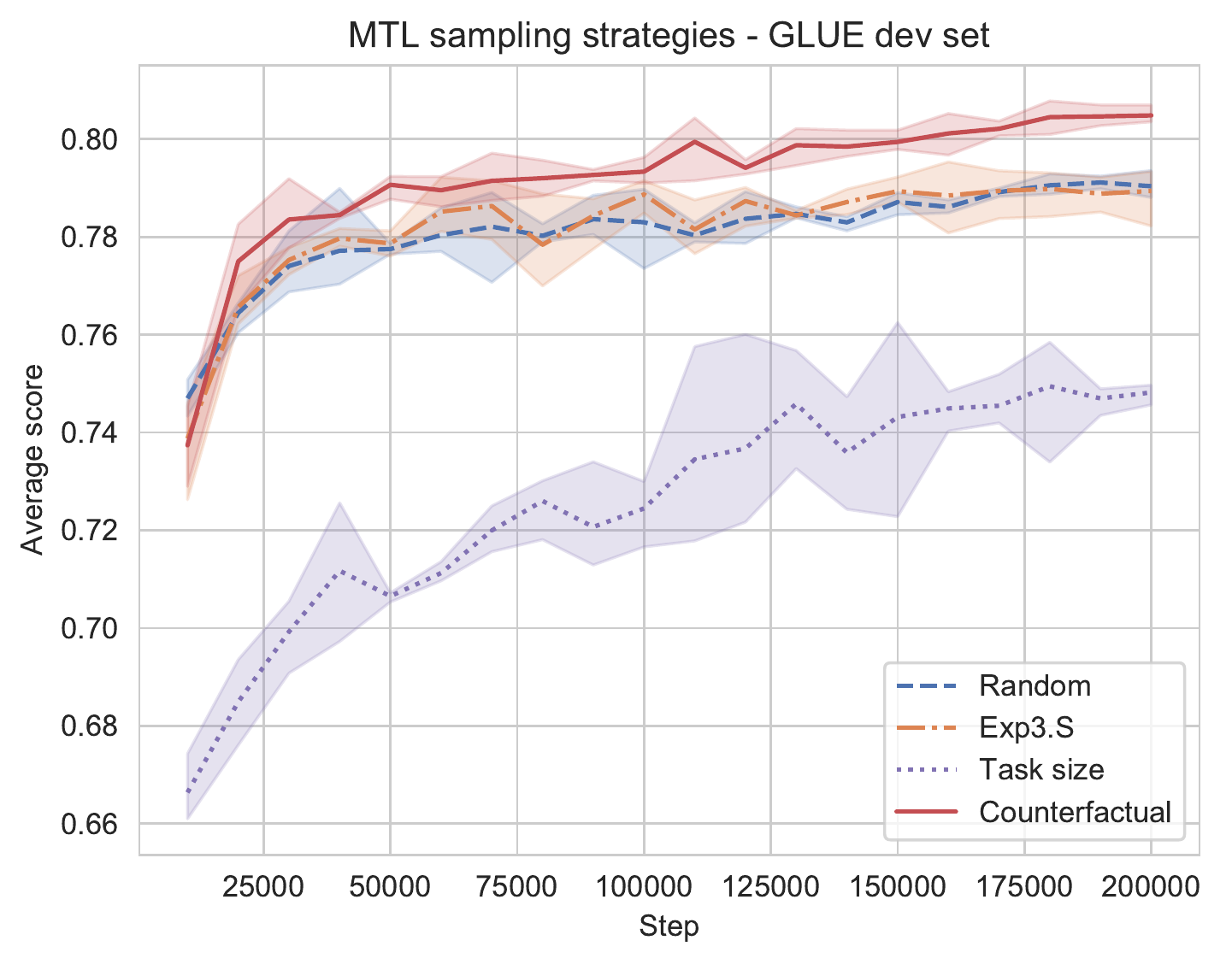}
\caption{Average dev set score across 8 GLUE tasks (excluding WNLI) over time for each policy.
Each line is the median value over 3 seeds, with shaded regions encompassing all seeds.}\label{fig:glue}
\end{figure}

Figure~\ref{fig:glue} shows average scores on the GLUE dev set over time for each policy, with the final
scores for each individual task given in Table~\ref{table:glue-dev}.
As in the bandit experiment, we find that $\pi^\textsc{random}$ and $\piexp$ perform similarly on average,
with neither reaching the final performance of our counterfactual method $\pi^\textsc{c}$.
$\pi^\textsc{task size}$ performs much worse than all other methods, as the large differences in
dataset sizes cause smaller tasks like CoLA to be under-sampled by this policy.

\begin{table*}[t!]
\begin{center}
\begin{tabular}{cccccccccc}
\toprule
\textbf{Policy} & \textbf{Average} & \textbf{CoLA} & \textbf{MNLI} & \textbf{MRPC} & \textbf{QNLI} & \textbf{QQP} & \textbf{RTE} & \textbf{SST-2} & \textbf{STS-B} \\ 
\midrule
$\pi^\textsc{random}$ & 79.1 & 48.2 & 82.9 & 88.8 & 88.2 & 86.9 & 73.4 & 92.1 & 72.1 \\
$\pi^\textsc{task size}$ & 74.9 & 18.5 & 83.6 & 86.1 & 77.4 & 88.6 & 77.2 & 90.6 & 76.6 \\
$\piexp$ & 78.9 & 49.2 & 83.6 & 87.6 & 87.3 & 83.7 & 74.5 & 91.8 & 73.8 \\
$\pic$ & 80.5 & 50.1 & 83.8 & 90.0 & 88.0 & 87.0 & 77.5 & 92.0 & 75.6 \\
\bottomrule
\end{tabular}
\end{center}
\caption{Individual task scores on the GLUE dev set for each task selection policy.
For tasks with multiple metrics, the task score is the average of their values, and we show the
average task score across all 3 random seeds.}\label{table:glue-dev}
\end{table*}

\begin{table}[t!]
\begin{center}
\begin{tabular}{lcc}
\toprule
\textbf{Task} & \textbf{Single task} & \textbf{MTL $\pi^\textsc{c}$} \\ 
\midrule
CoLA & 52.1 & 48.5 \\
MNLI (m/mm) & 84.6/83.4 &  83.5/83.1 \\
MRPC & 88.9/84.8 & 88.0/83.7 \\
QNLI & 90.5 & 90.5 \\
QQP & 71.2/89.2 & 70.4/88.7 \\
RTE & 66.4 & 74.5 \\
SST-2 & 93.5 & 93.1 \\
STS-B & 87.1/85.1 & 80.7/80.6 \\
WNLI & 65.1 & 65.1 \\
\midrule
GLUE Score & 78.3 & 77.9 \\
\bottomrule
\end{tabular}
\end{center}
\caption{GLUE test set results, comparing the single-task fine-tuning results for BERT with MTL fine-tuning using our learned policy $\pi^\textsc{c}$.}\label{table:glue}
\end{table}

We evaluated the best performing model using $\pi^\textsc{c}$ on the GLUE test set to make a better
comparison with single-task fine-tuning of $\text{BERT}_\textsc{base}$, and give the results in
Table~\ref{table:glue}.
The MTL model comes close to the GLUE score of the single-task fine-tuned models, albeit with some
differences in the individual task performances.
The MTL model is noticeably worse on CoLA, MNLI matched and STS-B, considerably better on RTE, and
similar on the remaining tasks.

\subsection{Discussion}\label{sec:discussion}
Our experiments show that choosing an appropriate task selection policy can have a large impact on
MTL performance, as highlighted by the gap between the best and worst polices on the GLUE dataset.
We see large performance gains for MTL on the RTE task in particular as reported in previous
work~\cite{wang2018a,devlin2018}, however the introduction of our task selection policy is not
sufficient to prevent some reduction in performance on CoLA, MNLI matched and STS-B.
We also note that a uniform random policy is a strong baseline in both of our experimental
settings, which confirms findings (on a different set of experiments) in~\citealp{graves2017}.

One weakness with our counterfactual method is the need to weight the estimation of the
target policy with a regularising entropy term, requiring an additional hyperparameter that
must be tuned. We believe that this is largely due to a deficiency in our surrogate sample-level reward
definition, perhaps this would not be necessary if we could devise a reward signal that aligns better
with the global MTL objective.
However, we are able to learn an improved policy on the GLUE benchmark with a relatively low number of training runs
(4 in total, including parameter tuning) --- one initial
run with a uniform random policy followed by 3 to pick the entropy weight.

The policy learned by our method on the GLUE dataset is shown in Table~\ref{table:policy}.
In general we see that the tasks with larger datasets are sampled more frequently, but
that it is important to boost the relative probability of sampling from tasks with smaller
datasets to maintain performance on them.
Task size alone is not completely indicative of
sampling frequency for our learned policy, as evidenced by STS-B and SST-2 being weighted similarly despite an
order of magnitude difference in their respective training set sizes, and the difference in weight between
MNLI and QQP which have similar training set sizes.
Our results support previous findings that it is often beneficial in MTL to spend more time on difficult or
larger tasks (a strategy that is sometimes referred to as ``anti-curriculum)~\cite{mccann2018,hokamp2019},
while suggesting a means for learning how to weight the tasks automatically.

\begin{table}[t!]
\begin{center}
\begin{tabular}{lcr}
\toprule
\textbf{Task} & \textbf{$\pi^\textsc{c}(\text{Task})$} & \textbf{$|D^\text{Task}|$} \\
\midrule
CoLA  & 0.089 & 10k \\
MNLI  & 0.255 & 393k \\
MRPC  & 0.086 & 4k \\
QNLI  & 0.134 & 108k \\
QQP   & 0.154 & 400k \\
RTE   & 0.086 & 2.7k \\
SST-2 & 0.094 & 67k \\
STS-B & 0.102 &  7k \\
\bottomrule
\end{tabular}
\end{center}
\caption{The fixed stochastic policy learned on the GLUE tasks using our counterfactual method
(rounded to 3 decimal places), and the size of the respective training sets.}\label{table:policy}
\end{table}


\section{Conclusion}
We evaluated several approaches to creating task selection policies for multitask learning,
and highlighted that in the context of the GLUE benchmark, the choice of policy can have a
large effect on overall performance.
We showed how the problem of task selection is related to the areas of curriculum learning
and off-policy evaluation, and suggested an approach based on counterfactual estimation
that leads to improved performance on a synthetic bandit-style task, as well as on the
more challenging GLUE tasks.
Interesting possibilities for future work include extending our learned policies to be
dynamic (instead of fixed stochastic),
evaluating additional counterfactual estimators~\cite{nedelec2017},
and devising (or learning) improved sample-level reward signals.

\bibliography{paper}
\bibliographystyle{acl_natbib}
\end{document}